\pdfoutput=1

\documentclass[11pt]{article}

\usepackage[final]{acl}

\usepackage{times}
\usepackage{latexsym}

\usepackage[T1]{fontenc}

\usepackage[utf8]{inputenc}

\usepackage{microtype}

\usepackage{inconsolata}

\usepackage{graphicx}
\usepackage{kotex}
\usepackage{bm}
\usepackage{amsmath}
\usepackage{graphicx}
\usepackage{amssymb}
\usepackage{booktabs}
\usepackage{colortbl}
\usepackage{multirow}
\usepackage{xcolor}
\usepackage{textcomp}  
\usepackage[most]{tcolorbox}
\usepackage{hyperref}
\usepackage{algorithm}
\usepackage{algorithmicx}
\usepackage{subcaption}
\usepackage{mathtools}
\usepackage{subcaption} 
\usepackage{enumitem}
\usepackage{algpseudocode}

\title{ToDi: Token-wise Distillation via Fine-Grained Divergence Control}

\author{
 \textbf{Seongryong Jung\textsuperscript{1,2}},
 \textbf{Suwan Yoon\textsuperscript{1}},
 \textbf{DongGeon Kim\textsuperscript{1}},
 \textbf{Hwanhee Lee\textsuperscript{1}\thanks{Corresponding author.}}
\\
 \textsuperscript{1}Department of Artificial Intelligence, Chung-Ang University,
 \textsuperscript{2}Dmtlabs
\\
 {
   \texttt{\{jungsr1116, swyoon0312, golddonggun, hwanheelee\}@cau.ac.kr}
 }
}

\begin{document}
\maketitle
\begin{abstract}

Large language models (LLMs) offer impressive performance but are impractical for resource-constrained deployment due to high latency and energy consumption. 
Knowledge distillation (KD) addresses this by transferring knowledge from a large teacher to a smaller student model. 
However, conventional KD, notably approaches like Forward KL (FKL) and Reverse KL (RKL), apply uniform divergence loss across the entire vocabulary, neglecting token-level prediction discrepancies. 
By investigating these representative divergences via gradient analysis, we reveal that FKL boosts underestimated tokens, while RKL suppresses overestimated ones, showing their complementary roles.
Based on this observation, we propose \textbf{Token-wise Distillation (ToDi)}, a novel method that adaptively combines FKL and RKL per token using a sigmoid-based weighting function derived from the teacher-student probability log-ratio. 
ToDi dynamically emphasizes the appropriate divergence for each token, enabling precise distribution alignment. 
We demonstrate that ToDi consistently outperforms recent distillation baselines using uniform or less granular strategies across instruction-following benchmarks. Extensive ablation studies and efficiency analysis further validate ToDi's effectiveness and practicality.\footnote{The code is available at \url{https://github.com/jungseongryong/ToDi}}

\end{abstract}

\section{Introduction}

Recent advances in large language models (LLMs), driven by scaling up model size, have substantially enhanced their ability to follow user instructions and generate contextually appropriate responses \citep{NEURIPS2020_1457c0d6, sanh2022multitask, wei2022finetuned, chung2024scaling}.  
However, the continued enlargement of model size introduces several challenges, including increased inference latency, high energy consumption, and inefficiency in resource-constrained environments.  
To address these issues, knowledge distillation (KD; \citealp{hinton2015distilling}) has been widely adopted; this approach aims to minimize the performance gap between teacher and student models by transferring knowledge from a high-performing large teacher model to a smaller student model. Recently, various knowledge distillation techniques for enhancing the efficiency of LLMs have been proposed, and research surrounding these methods is actively underway (\citealp{zhang-etal-2024-dual}; \citealp{feng-etal-2024-teaching-small}; \citealp{shing2025taid}).

\begin{figure}[t]
  \centering
  \includegraphics[width=\columnwidth]{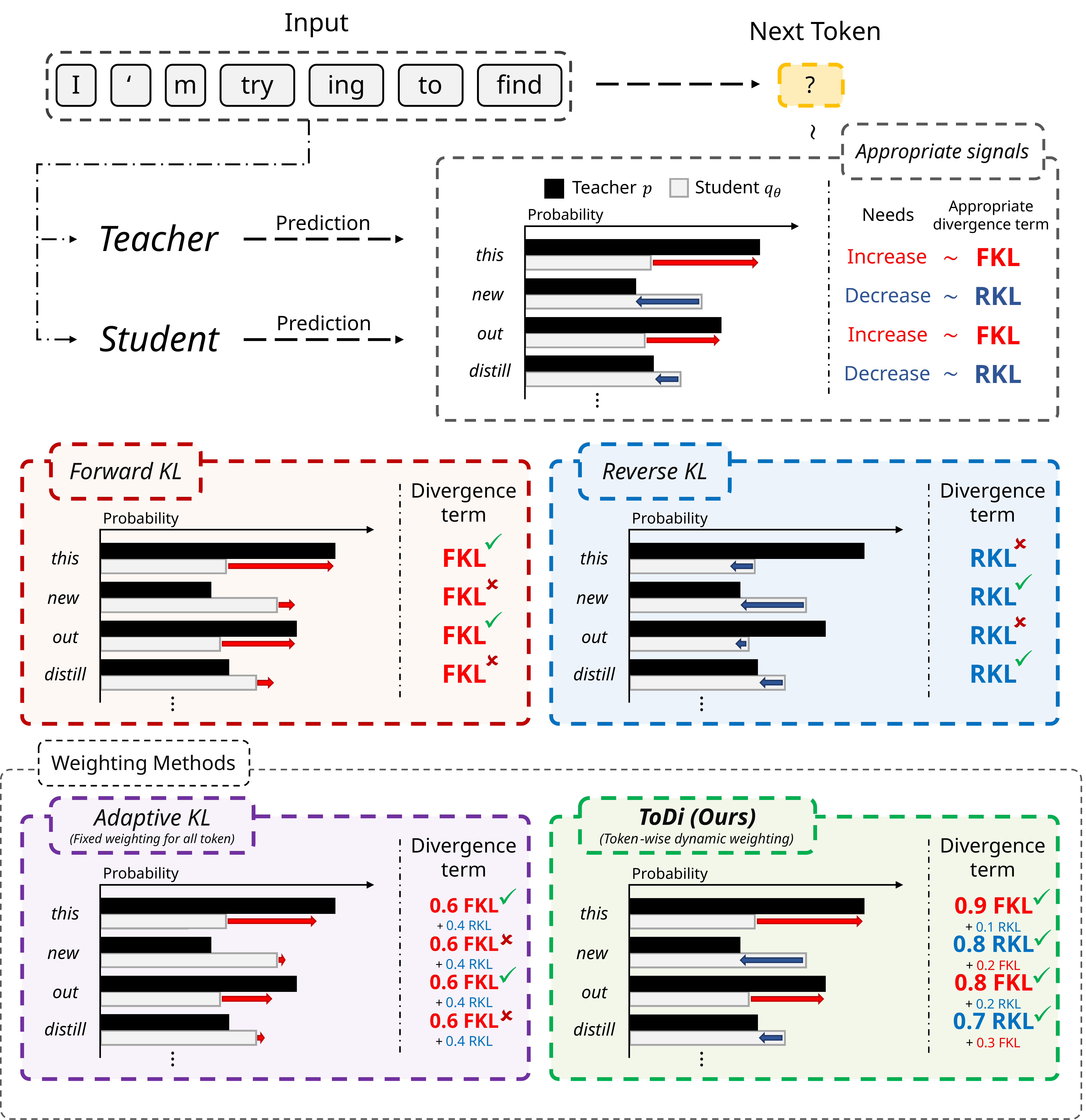}
  \caption{Token-wise learning signals for KL-based distillation objectives. Conventional methods apply a fixed divergence across the entire vocabulary, while ToDi dynamically blends Forward and Reverse KL per-token based on the teacher–student probability ratio, balancing gradients across all tokens.}
  \label{fig:intro}
  \vspace{-5mm}
\end{figure}

Conventional knowledge distillation methods often employ divergences such as Forward KL (FKL) and Reverse KL (RKL) to minimize the discrepancy between teacher and student distributions~\citep{hinton2015distilling, gu2024minillm}.
However, as depicted in the example for FKL and RKL in Figure~\ref{fig:intro}, these approaches apply a single divergence uniformly across the \emph{entire vocabulary}, regardless of how severely the student misestimates each token.
This uniform-loss assumption persists in symmetric and hybrid variants of FKL/RKL~\citep{wen-etal-2023-f, ko2024distillm, agarwal2024policy}, and even dynamic combinations at the vocabulary-set or time-step level like Adaptive KL~\citep{wu-etal-2025-rethinking}.
We hypothesize that such uniform treatment is sub-optimal because different tokens may require different correction signals.

In this paper, we analyze the limitation of uniform application by investigating token-specific optimal signals through a gradient-based analysis of divergences in existing KD methods (Section~\ref{sec:grad_behavior}). This analysis reveals that FKL effectively increases the probability of tokens that the student model underestimates relative to the teacher model, whereas RKL excels at suppressing the probability of tokens that it overestimates, showing their distinct and complementary roles. 
However, existing methods apply a uniform divergence loss across the entire vocabulary, failing to leverage these complementary signals effectively at the token level. As shown in Figure~\ref{fig:intro}, this uniformity prevents appropriate training signals for individual tokens, particularly when the student significantly over- or underestimates the teacher's distribution.

Motivated by this insight, we propose a novel distillation method, \textbf{Token-wise Distillation (ToDi)} (Section~\ref{sec:ToDi}). As illustrated in Figure~\ref{fig:intro}, ToDi dynamically balances the contributions of FKL and RKL based on token-level prediction discrepancies by adaptively combining them per-token using a token-specific weighting function. This approach directly provides tailored training signals that capture fine-grained differences between the teacher and student distributions, going beyond uniform loss application.

We demonstrate ToDi's effectiveness through extensive experiments and show that ToDi consistently outperforms recent distillation baselines on various instruction-following benchmarks, achieving superior ROUGE-L scores and higher win rates in GPT-4-based pairwise evaluations. Furthermore, we validate the critical importance of ToDi's token-wise divergence control. We also show that ToDi maintains stable training and linear time complexity with respect to vocabulary size, highlighting its efficiency and practicality.

The principal contributions of this paper are as follows:
\begin{itemize}[leftmargin=10pt, labelindent=0pt]

\item  We analyze and show the complementary roles of FKL and RKL for KD through gradient analysis.

\item Based on this analysis, we propose ToDi, a new KD method that adaptively combines FKL and RKL per token according to prediction discrepancies and enables fine-grained distribution alignment.

\item We provide theoretical grounding for ToDi and demonstrate its superior performance over existing methods through extensive experiments on instruction following tasks.
\end{itemize}

\section{Related Work}

\subsection{Objective Functions of KD}
In knowledge distillation (\citealp{hinton2015distilling}), the student model is trained to mimic the teacher’s output distribution by minimizing the divergence loss.
The FKL induces mode averaging, smoothing a multimodal teacher distribution, while the RKL causes mode collapse, driving the student to focus on a single mode (\citealp{koller2009probabilistic,chan2022greedification,wang2024beyond}).  
To counter these extremes, \citet{wen-etal-2023-f} adopted the symmetric Jensen–Shannon Divergence (JSD), and \citet{agarwal2024policy} generalized it to interpolate between FKL and RKL. Skewed KL variants (SKL, SRKL) further mix the student distribution into the teacher’s distribution for stability~\citep{ko2024distillm}, while TAID~\citep{shing2025taid} inserts a time-varying intermediate distribution between teacher and student. 

Despite these advances, all prior work on applying KD for language models still processes the \emph{entire} vocabulary distribution at every sequence position and applies a uniform loss across tokens. This coarse treatment misses token-level mismatches between teacher and student, limiting the student’s ability to replicate the teacher’s fine-grained predictive structure. Our proposed method aims to overcome this limitation by applying a token-wise dynamic divergence control, precisely addressing these fine-grained mismatches.

\subsection{Dynamic Combination of FKL and RKL}
Several studies have explored combining FKL and RKL to take advantage of both methods. \citet{lee2023self} proposed a straightforward additive combination, whereas \citet{amara2022bd} introduced BD-KD, which adjusts the weights of FKL and RKL on a per-sample basis via the entropy gap between teacher and student distributions. \citet{wu-etal-2025-rethinking} presented AKL—tailored for LLM distillation—that adaptively combines the two divergences based on the observation that, in early training, FKL primarily learns head predictions while RKL focuses on tail predictions.
More recently, \citet{ko2025distillm} proposed DistiLLM-2, a contrastive framework that applies distinct divergence functions depending on whether the responses are generated by the teacher or the student.

Nevertheless, such approaches still dynamically apply FKL and RKL to the entire vocabulary distribution at every sequence position without assigning \emph{dynamic weights to individual tokens}. This limitation prevents a fine-grained reflection of token-level prediction differences between teacher and student, thereby hindering the learning of detailed predictive structures. In contrast, our proposed ToDi method dynamically balances FKL and RKL on a per‐token basis, capturing fine‐grained probability discrepancies and enabling more precise predictive structure learning.

\section{Gradient Behavior of FKL and RKL}
\label{sec:grad_behavior}

In this section, we formalize knowledge distillation for autoregressive LLMs and analyze the FKL and RKL objectives from a gradient perspective. 
By understanding the gradients, we precisely examine how the learning signal for each vocabulary token depends on the relative magnitudes of the teacher probability \(p(v_i \mid \mathbf{y}_{<t}, \mathbf{x})\) and the student probability \(q_\theta(v_i \mid \mathbf{y}_{<t}, \mathbf{x})\), providing insight into token-specific optimal signals.

\subsection{Preliminaries}
\label{subsec:preliminaries}

An autoregressive LLMs generates an output sequence \(\mathbf{y} = [y_1,\dots,y_{|\mathbf{y}|}]\) conditioned on an input sequence \(\mathbf{x}\). At each time step \(t\), it selects one token from a finite vocabulary \(\mathcal{V} = \{v_1,\dots,v_{|\mathcal{V}|}\}\). 

KD minimizes the discrepancy between the teacher’s distribution \(p(y_t \mid \mathbf{y}_{<t}, \mathbf{x})\) and the student’s distribution \(q_\theta(y_t \mid \mathbf{y}_{<t}, \mathbf{x})\), where \(\theta\) denotes the student parameters and \(\mathbf{y}_{<t} = [y_1, \dots, y_{t-1}]\) are the tokens generated before step \(t\).

During KD, the loss is typically instantiated as either the \textit{FKL} or the \textit{RKL}. At time step \( t \), the contribution of each divergence for a token \( v_i \in \mathcal{V} \) is defined as:
\begin{equation}
D_{\text{FKL}}^{(t,i)}(p, q_{\theta}) = 
p\!\left( v_i \,\middle|\, \mathbf{y}_{<t}, \mathbf{x} \right) 
\log \frac{
p\!\left( v_i \,\middle|\, \mathbf{y}_{<t}, \mathbf{x} \right)
}{
q_{\theta}\!\left( v_i \,\middle|\, \mathbf{y}_{<t}, \mathbf{x} \right)
},
\label{eq:fkl_def}
\end{equation}
\begin{equation}
D_{\text{RKL}}^{(t,i)}(p, q_{\theta}) = 
q_{\theta}\!\left( v_i \,\middle|\, \mathbf{y}_{<t}, \mathbf{x} \right) 
\log \frac{
q_{\theta}\!\left( v_i \,\middle|\, \mathbf{y}_{<t}, \mathbf{x} \right)
}{
p\!\left( v_i \,\middle|\, \mathbf{y}_{<t}, \mathbf{x} \right)
}.
\label{eq:rkl_def}
\end{equation}

\paragraph{Training Objective}
We accumulate the token-level divergences (from Equations~\ref{eq:fkl_def} and \ref{eq:rkl_def}) over all time steps and vocabulary entries to obtain the total forward and reverse KL divergence losses:
\begin{align}
\mathcal{L}_{\mathrm{FKL}}
&=
\sum_{t=1}^{|\mathbf{y}|}\;\sum_{i=1}^{|\mathcal{V}|}
   D_{\mathrm{FKL}}^{(t,i)}\bigl(p,\,q_\theta\bigr),
   \label{eq:fkl_total}\\[4pt]
\mathcal{L}_{\mathrm{RKL}}
&=
\sum_{t=1}^{|\mathbf{y}|}\;\sum_{i=1}^{|\mathcal{V}|}
   D_{\mathrm{RKL}}^{(t,i)}\bigl(p,\,q_\theta\bigr).
   \label{eq:rkl_total}
\end{align}



\subsection{Theoretical Analysis}
\label{subsec:analysis_kl}

We theoretically analyze the FKL and RKL training signals.  
In particular, we examine how the two divergences exert opposite corrective effects depending on the relative magnitudes of the teacher distribution \(p(y_t \mid \mathbf{y}_{<t}, \mathbf{x})\) and the student distribution \(q_{\theta}(y_t \mid \mathbf{y}_{<t}, \mathbf{x})\).  
The analysis is grounded in the token-level definitions given in Equations~\ref{eq:fkl_def} and~\ref{eq:rkl_def}.

\begin{figure*}[!ht]
  \centering
  \includegraphics[width=\linewidth]{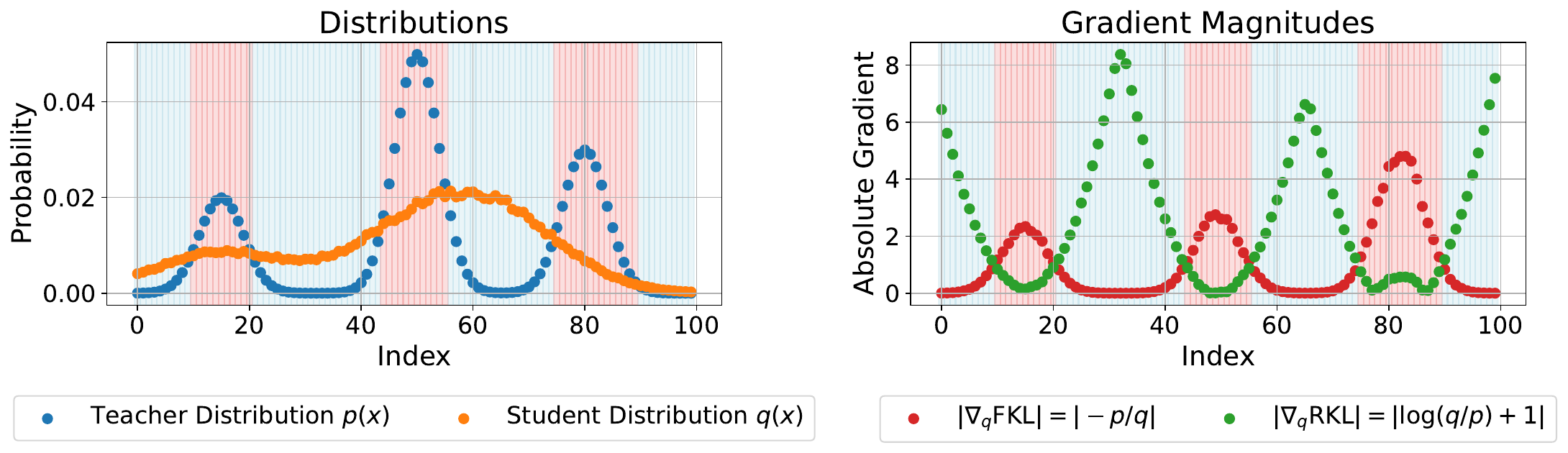}
  \caption{Toy example demonstrating the behavior of FKL and RKL gradients. In regions where \( p > q \), FKL provides stronger gradients, while in regions where \( q > p \), RKL provides stronger learning signals.}

    \label{fig:toy_gradients}
    \vspace{-2mm}
\end{figure*}

\paragraph{Gradient Form.}
The partial derivatives of each divergence with respect to \(q_{\theta}\) are:

\begingroup
    \small
\begin{equation}
\frac{\partial}{\partial q_\theta(v_i \mid \mathbf{y}_{<t}, \mathbf{x})}
D_{\text{FKL}}^{(t,i)}(p, q_\theta)
= -\frac{p(v_i \mid \mathbf{y}_{<t}, \mathbf{x})}
       {q_\theta(v_i \mid \mathbf{y}_{<t}, \mathbf{x})},
\label{eq:fkl_deriv}
\end{equation}
\endgroup

\begingroup
    \small
\begin{equation}
\begin{aligned}
\frac{\partial}{\partial q_\theta(v_i \mid \mathbf{y}_{<t}, \mathbf{x})}
D_{\text{RKL}}^{(t,i)}(p, q_\theta)
&= \log
   \frac{q_\theta(v_i \mid \mathbf{y}_{<t}, \mathbf{x})}
        {p(v_i \mid \mathbf{y}_{<t}, \mathbf{x})}+ 1.
\end{aligned}
\label{eq:rkl_deriv}
\end{equation}
\endgroup

We describe the detailed derivations in Appendix~\ref{sec:appendix_gradient}.

\paragraph{Difference in Training Signals by Relative Probability.}
The two gradients can be compared through a single ratio
\(r = \frac{p\bigl(v_i \mid \mathbf{y}_{<t}, \mathbf{x}\bigr)}
         {q_\theta\bigl(v_i \mid \mathbf{y}_{<t}, \mathbf{x}\bigr)}\):
\begin{itemize}
    \item \textbf{\(r>1\) (the student model underestimates).}  
    Here, the FKL gradient \(-r\) is a negative value whose magnitude exceeds~1, pushing \(q_\theta\) to \textit{increase} sharply.  
    The RKL gradient, \(\log \tfrac{1}{r}+1\), turns negative only when \(r>e\) and its magnitude is smaller, producing a relatively weak corrective signal.  
    Thus, for tokens underestimated by the student, FKL provides the dominant "push-up" signal.

    \item \textbf{\(r<1\) (the student model overestimates).}  
    In this case, the FKL gradient remains a small negative value, whereas the RKL gradient is a positive value greater than~1, providing a strong signal to \textit{decrease} \(q_\theta\).  
    Consequently, when the student overestimates, RKL provides the dominant "pull-down" signal.
\end{itemize}

\begin{table}[h]
  \centering
  \small
  \renewcommand{\arraystretch}{1.7}
  \begin{tabular}{c|cc}
    \toprule
    \textbf{Case} & \textbf{Forward KL} & \textbf{Reverse KL} \\
    \midrule
    \(\mathbf{p > q_\theta}\) & 
      \(\boldsymbol{\uparrow}\) \textbf{Strong push-up} & 
      \(\approx\) \emph{Weak push-up} \\
    \(\mathbf{p < q_\theta}\) & 
      \(\approx\) \emph{Weak pull-down} & 
      \(\boldsymbol{\downarrow}\) \textbf{Strong pull-down} \\
    \bottomrule
  \end{tabular}
  \caption{Complementary training signals of FKL vs. RKL.}
  \label{tab:kl_gradient_summary}
\end{table}

In summary, as organized in Table~\ref{tab:kl_gradient_summary}, our theoretical analysis reveals that FKL and RKL provide complementary training signals around the boundary \(r=1\): FKL strongly encourages increasing student probability (i.e. push-up) for underestimated tokens (\(p>q_\theta\)), while RKL strongly encourages decreasing student probability (i.e. pull-down) for overestimated tokens (\(q_\theta>p\)).

\subsection{Empirical Analysis of a Toy Example}
\label{subsec:empirical}
To empirically examine how FKL and RKL gradient magnitudes depend on the relative teacher–student probabilities at each token, we construct a toy example by defining teacher distribution \(p(x)\) and student distribution \(q(x)\).
Figure~\ref{fig:toy_gradients} illustrates the comparison of gradient magnitudes according to the relative relationship between the teacher distribution \(p(x)\) and the student distribution \(q(x)\) in a toy example. 
The left panel shows where the two distributions intersect, with the regions \(p(x)>q(x)\) and \(q(x)>p(x)\) shaded separately. The right panel visualizes, for each index, the gradient magnitudes induced by FKL and RKL.

Consistent with the theoretical analysis, we observe in the toy example that in the region where \(p(x) > q(x)\), FKL produces substantially larger gradients than RKL, delivering a strong corrective signal for tokens that the student under-estimates relative to the teacher. Conversely, in the region where \(q(x) > p(x)\), the magnitude of the RKL gradient is greater, indicating a strong signal to suppress over-estimation. Consequently, FKL and RKL provide specialized training signals in different scenarios.

\begin{figure*}[t]
  \centering
  \captionsetup{skip=1pt}
  \includegraphics[width=1\linewidth]{ToDi.pdf}
  \caption{Illustration of the Token‐wise Distillation. \textbf{(Left)} For each vocabulary token, the contributions of FKL and RKL are dynamically combined using a token‐specific weight $\alpha_{t,i}$.  \textbf{(Right)} The weight $\alpha_{t,i}$, determined by the teacher–student probability ratio, smoothly increases FKL emphasis when $p > q_\theta$ and RKL emphasis when $q_\theta > p$.}
    \label{fig:method}
    \vspace{-1.5mm}
\end{figure*}

\section{ToDi}
\label{sec:ToDi}

In this section, we introduce \textbf{Token-wise Distillation (ToDi)}, which dynamically adjusts the contributions of FKL and RKL based on the token‐level probability ratios in the teacher and student distributions.

\paragraph{Objective Functions for ToDi.}
As shown in the gradient analysis of Section~\ref{sec:grad_behavior}, for each vocabulary token \(v_i\), when 
\(
p(v_i \mid \mathbf{y}_{<t}, \mathbf{x}) > q_\theta(v_i \mid \mathbf{y}_{<t}, \mathbf{x}),
\)
the FKL provides a learning signal that effectively increases \(q_\theta\), and conversely, when
\(
q_\theta(v_i \mid \mathbf{y}_{<t}, \mathbf{x}) > p(v_i \mid \mathbf{y}_{<t}, \mathbf{x}),
\)
the RKL offers a signal that reduces \(q_\theta\). 
Building on this insight into their complementary roles, we propose a novel distillation method, \textbf{Token‐wise Distillation (ToDi)}, which dynamically combines FKL and RKL according to the relative magnitudes of the teacher probability \(p(v_i \mid \mathbf{y}_{<t}, \mathbf{x})\) and the student probability \(q_\theta(v_i \mid \mathbf{y}_{<t}, \mathbf{x})\). 
Unlike conventional approaches that apply a single loss uniformly across the entire vocabulary, ToDi computes a specific loss for each token \(v_i\) at time step \(t\), denoted \(D_{\text{ToDi}}^{(t,i)}\).
This token-level loss is a weighted sum of the token's FKL and RKL divergences.
Specifically, the token-level loss \(D_{\text{ToDi}}^{(t,i)}\) is defined as follows:

\begin{equation}
\begin{aligned}
\label{eq:ToDi}
D_{\text{ToDi}}^{(t,i)}(p, q_\theta) = {} & \alpha_{t,i}\,\! \cdot D_{\text{FKL}}^{(t,i)}(p, q_\theta) \\
& + \bigl(1 - \alpha_{t,i}\bigr)\, \cdot D_{\text{RKL}}^{(t,i)}(p, q_\theta),
\end{aligned}
\end{equation}
where \(\alpha_{t,i}\) is a \textbf{token-specific weight} dynamically computed for each token \(v_i\) based on the relative teacher and student probabilities. 

As illustrated in Figure~\ref{fig:method} (Left), we utilize the weighting function to amplify the contribution of FKL when needed (when \(p > q_\theta\)) and amplify the contribution of RKL when needed (when \(q_\theta > p\)).

The overall distillation loss is then the sum of these token-level losses over all time steps and vocabulary entries:
\begin{equation}
\mathcal{L}_{\text{ToDi}} = \sum_{t=1}^{|\mathbf{y}|} \sum_{i=1}^{|\mathcal{V}|} D_{\text{ToDi}}^{(t,i)}(p, q_\theta).
\end{equation}

\paragraph{Weighting Function for ToDi.} 
The core of ToDi's token-wise control lies in the weighting function that determines \(\alpha_{t,i}\). This weight must dynamically adjust according to the relative magnitudes of \(p(v_i \mid \mathbf{y}_{<t}, \mathbf{x})\) and \(q_\theta(v_i \mid \mathbf{y}_{<t}, \mathbf{x})\) to effectively leverage the complementary nature of FKL and RKL. 

Specifically, the token-specific weight \(\alpha_{t,i}\) is defined by a function \(\textit{W}\) of these probabilities:
\begin{equation}
\alpha_{t,i} = \textit{W}\bigl(p(v_i \mid \mathbf{y}_{<t}, \mathbf{x}),\, q_\theta(v_i \mid \mathbf{y}_{<t}, \mathbf{x})\bigr)
\end{equation}

The function \(\textit{W}\) should assign a larger value (thus increasing the contribution of FKL) when 
\(
p(v_i \mid \mathbf{y}_{<t}, \mathbf{x}) > q_\theta(v_i \mid \mathbf{y}_{<t}, \mathbf{x})
\),
so as to boost the student’s probability. Conversely, when 
\(
q_\theta(v_i \mid \mathbf{y}_{<t}, \mathbf{x}) > p(v_i \mid \mathbf{y}_{<t}, \mathbf{x}),
\)
a smaller function value (favoring RKL) is appropriate. 
To satisfy these requirements and enable fine-grained control, the function \textit{W} must meet the following four conditions:
\begin{itemize}
    \item If \(p(v_i \mid \mathbf{y}_{<t}, \mathbf{x}) > q_\theta(v_i \mid \mathbf{y}_{<t}, \mathbf{x})\), then \(\alpha_{t,i}\) should be greater than 0.5 to emphasize FKL.
    \item If \(q_\theta(v_i \mid \mathbf{y}_{<t}, \mathbf{x}) > p(v_i \mid \mathbf{y}_{<t}, \mathbf{x})\), then \(\alpha_{t,i}\) should be less than 0.5 to emphasize RKL.
    \item To allocate more extreme weights when the teacher–student probability gap is larger, \(\alpha_{t,i}\) must be a monotonically increasing function of the ratio 
    \(\displaystyle p(v_i \mid \mathbf{y}_{<t}, \mathbf{x})/{q_\theta(v_i \mid \mathbf{y}_{<t}, \mathbf{x})}\).
    \item \(\alpha_{t,i}\) must lie within the valid weight range \([0,1]\).
\end{itemize}

To satisfy all four conditions, we adopt the sigmoid function for \(\textit{W}\), defining \(\alpha_{t,i}\) as:
\begin{equation}
\alpha_{t,i}
= \mathrm{sg} \biggr[\sigma\!\Biggl(
\log \frac{p\bigl(v_i \mid \mathbf{y}_{<t}, \mathbf{x}\bigr)}
         {q_\theta\bigl(v_i \mid \mathbf{y}_{<t}, \mathbf{x}\bigr)}
\Biggr)\biggr]
\end{equation}

Here, $\sigma(\cdot)$ denotes the sigmoid function, and $\mathrm{sg}[\cdot]$ the stop‐gradient operator. 
By applying $\mathrm{sg}[\cdot]$, we block gradient flow through its arguments, effectively treating the weight \(\alpha_{t,i}\) as a fixed value during the backpropagation of the loss.

As illustrated in Figure~\ref{fig:method} (Right), \(\alpha_{t,i}\) smoothly varies between 0 and 1 according to the magnitude of 
\(\displaystyle p(v_i \mid \mathbf{y}_{<t}, \mathbf{x})/{q_\theta(v_i \mid \mathbf{y}_{<t}, \mathbf{x})}\),
naturally reflecting the teacher–student probability discrepancy. A detailed proof that the sigmoid satisfies all four conditions is provided in Appendix~\ref{app:sigmoid}.
Furthermore, we implement the stop‐gradient operator \(\mathrm{sg}[\cdot]\) as a \texttt{detach} operation during training; its effects are discussed in detail in Appendix~\ref{app:todi_vs_jeffreys}.

\begin{table}[!h]
  \centering
  \small
  \begin{tabular}{@{}lcc@{}}
    \toprule
    \textbf{Function} & $\alpha_{t,i}(r)\;(r=p/q_\theta)$ & $\beta$ \\
    \midrule
    Sigmoid            & $\tfrac{1}{1+e^{-\log r}} = \tfrac{r}{1+r}$ & $1$ \\
    Scaled tanh        & $\displaystyle\tfrac12\bigl(1+\tanh(\log r)\bigr)$                      & $2$ \\
    Jeffreys (fixed)   & $\displaystyle\tfrac12$                                                & $0$ \\
    Step function      & $\displaystyle\mathbf{1}[\,r>1\,]$                                     & $\beta\!\to\!\infty$ \\
    \bottomrule
  \end{tabular}
  \caption{Various weighting functions can be unified under the Generalized ToDi, where each can be expressed in the form $\alpha_{t,i}(r) = \sigma(\beta \log r)$ with an appropriate scaling factor $\beta$.}
  \label{tab:gtd_family}
  \vspace{-3mm}
\end{table}

\paragraph{Generalized ToDi.}

Any function satisfying the four weight conditions introduced above can take many forms.
To explore this design space and unify various weighting strategies, we introduce a \emph{scaling hyperparameter} \(\beta \in \mathbb{R}\). By incorporating \(\beta\) into the sigmoid input, we can express a variety of weighting functions in a \emph{single unified form}. In this case, the ToDi weight function \(\alpha_{t,i}\) is defined as:

\begin{equation}
\alpha_{t,i}
= \mathrm{sg} \biggr[\sigma\!\left(
\beta \cdot \log 
\frac{
p\bigl(v_i \mid \mathbf{y}_{<t}, \mathbf{x}\bigr)
}{
q_\theta\bigl(v_i \mid \mathbf{y}_{<t}, \mathbf{x}\bigr)
}
\right)\biggr]
\end{equation}

As summarized in Table~\ref{tab:gtd_family}, by simply varying the value of \(\beta\), this unified framework can represent a range of weighting functions, such as the standard sigmoid (\(\beta=1\)), scaled tanh (\(\beta=2\)), Jeffreys divergence~\citep{jeffreys1946invariant} (\(\beta=0\)), and approximating a step function (\(\beta \to \infty\)).

\section{Experiments}
\label{sec:Experiments}

\subsection{Experimental Setup}
\paragraph{Training Configuration.}  
We follow the experimental setup of \citet{zhang-etal-2024-dual} to evaluate ToDi. For training, we use the \texttt{databricks/dolly-15k} dataset, which comprises 11K training samples, 1K validation samples, and 500 test samples. As student models for the main experiment, we employ GPT2-120M \citep{radford2019language} and TinyLLaMA-1.1B \citep{zhang2024tinyllama}. We train GPT2-120M via full fine-tuning using GPT2-1.5B as the teacher model, whereas we train TinyLLaMA-1.1B with LoRA \citep{hu2022lora} using LLaMA2-7B \citep{touvron2023llama} as the teacher.

\paragraph{Evaluation Protocol.}  
We conduct performance evaluation following the protocol of \citet{gu2024minillm}, using the ROUGE-L metric \citep{lin-2004-rouge}.  We assess instruction-following ability across five datasets: \textbf{DollyEval}, \textbf{S-NI} \citep{wang-etal-2022-super}, \textbf{UnNI} \citep{honovich-etal-2023-unnatural}, \textbf{SelfInst} \citep{wang-etal-2023-self-instruct}, and \textbf{VicunaEval} \citep{zheng2023judging}. We repeat each evaluation with five different random seeds, and we report the average scores. Further details of the experimental setup are provided in Appendix~\ref{app:Experimental Details}.  

\paragraph{Baseline Methods.}
We use the following methods as baselines to compare the performance of ToDi:
\begin{itemize}
    \item \textbf{SFT}: Fine-tuning the student model directly on the dataset without knowledge distillation.
    \item \textbf{FKL/RKL} (\citealp{hinton2015distilling}; \citealp{gu2024minillm}): Knowledge distillation using Forward or Reverse KL divergence.
    \item \textbf{JS/TVD} (\citealp{wen-etal-2023-f}): Symmetric divergences—Jensen–Shannon and Total Variation—minimizing the distance between the teacher and student distributions.
    \item \textbf{SKL/SRKL} (\citealp{ko2024distillm}): Skewed KL and Skewed Reverse KL, which mix teacher and student distributions at ratio \(\lambda\); SKL uses \(\lambda p + (1-\lambda)q_\theta\) while SRKL uses \((1-\lambda)p + \lambda q_\theta\).
    \item \textbf{AKL} (\citealp{wu-etal-2025-rethinking}): Adaptive KL that combines FKL and RKL by considering head–tail differences in the distributions.
\end{itemize}

To evaluate ToDi’s performance, we select various divergence-based knowledge distillation methods as baselines and compare their performance based on the choice of divergence.

\begin{table*}[!ht]
\small
\centering
\makebox[\linewidth][c]{%
\begin{tabular}{llccccc|c}
\toprule
\multicolumn{2}{l}{\textbf{Methods}} & \textbf{DollyEval} & \textbf{S-NI} & \textbf{UnNI} & \textbf{SelfInst} & \textbf{VicunaEval} & \textbf{Average}  \\
\midrule
\midrule
\multicolumn{8}{c}{GPT2 1.5B → GPT2 120M} \\ 
\midrule
\rowcolor{gray!30}
\multicolumn{2}{l}{Teacher} & 26.66{\scriptsize\textpm0.30} & 27.17{\scriptsize\textpm0.33} & 31.60{\scriptsize\textpm0.13} & 14.42{\scriptsize\textpm0.49} & 16.32{\scriptsize\textpm0.41} & 23.23 \\
\multicolumn{2}{l}{SFT}     & 23.09{\scriptsize\textpm0.53} & 16.44{\scriptsize\textpm0.39} & 18.96{\scriptsize\textpm0.08} & 9.72{\scriptsize\textpm0.43} & 14.81{\scriptsize\textpm0.34} & 16.61 \\
\multicolumn{2}{l}{FKL}     & 24.06{\scriptsize\textpm0.43} & 18.43{\scriptsize\textpm0.22} & 21.42{\scriptsize\textpm0.04} & 11.13{\scriptsize\textpm0.34} & 15.53{\scriptsize\textpm0.45} & 18.12 \\
\multicolumn{2}{l}{RKL}     & 24.22{\scriptsize\textpm0.18} & \underline{18.60{\scriptsize\textpm0.10}} & \underline{21.99{\scriptsize\textpm0.07}} & \textbf{11.42{\scriptsize\textpm0.33}} & \textbf{15.65{\scriptsize\textpm0.51}} & \underline{18.38} \\
\multicolumn{2}{l}{JS}      & 23.77{\scriptsize\textpm0.29} & 17.31{\scriptsize\textpm0.17} & 19.74{\scriptsize\textpm0.07} & 10.08{\scriptsize\textpm0.37} & 15.08{\scriptsize\textpm0.32} & 17.20 \\
\multicolumn{2}{l}{TVD}     & 23.90{\scriptsize\textpm0.61} & 17.89{\scriptsize\textpm0.24} & 20.87{\scriptsize\textpm0.12} & 10.73{\scriptsize\textpm0.71} & 15.20{\scriptsize\textpm0.30} & 17.72 \\
\multicolumn{2}{l}{SKL}     & 24.05{\scriptsize\textpm0.31} & 17.18{\scriptsize\textpm0.31} & 20.43{\scriptsize\textpm0.08} & 10.54{\scriptsize\textpm0.55} & 14.93{\scriptsize\textpm0.29} & 17.42 \\
\multicolumn{2}{l}{SRKL}    & 24.20{\scriptsize\textpm0.40} & 18.02{\scriptsize\textpm0.18} & 21.67{\scriptsize\textpm0.09} & 11.05{\scriptsize\textpm0.48} & 15.07{\scriptsize\textpm0.22} & 18.00 \\
\multicolumn{2}{l}{AKL}     & \underline{24.67{\scriptsize\textpm0.29}} & 18.29{\scriptsize\textpm0.23} & 21.46{\scriptsize\textpm0.12} & 10.62{\scriptsize\textpm0.68} & 15.28{\scriptsize\textpm0.16} & 18.07 \\
\rowcolor{yellow!30}
\multicolumn{2}{l}{\textbf{ToDi (Ours)}}    & \textbf{24.81{\scriptsize\textpm0.62}} & \textbf{19.42{\scriptsize\textpm0.18}} & \textbf{22.16{\scriptsize\textpm0.21}} & \underline{11.30{\scriptsize\textpm0.41}} & \underline{15.61{\scriptsize\textpm0.34}} & \textbf{18.66} \\
\midrule
\midrule
\multicolumn{8}{c}{LLaMA2 7B → TinyLLaMA 1.1B} \\
\midrule
\rowcolor{gray!30}
\multicolumn{2}{l}{Teacher} & 28.88{\scriptsize\textpm0.23} & 30.72{\scriptsize\textpm0.36} & 32.02{\scriptsize\textpm0.08} & 19.89{\scriptsize\textpm0.58} & 18.76{\scriptsize\textpm0.59} & 26.05 \\
\multicolumn{2}{l}{SFT}     & 23.36{\scriptsize\textpm0.26} & 26.19{\scriptsize\textpm0.18} & 26.69{\scriptsize\textpm0.08} & 15.76{\scriptsize\textpm1.04} & 15.88{\scriptsize\textpm0.63} & 21.58 \\
\multicolumn{2}{l}{FKL}     & 25.40{\scriptsize\textpm0.50} & 30.13{\scriptsize\textpm0.43} & 29.47{\scriptsize\textpm0.06} & \textbf{18.22{\scriptsize\textpm1.12}} & 16.77{\scriptsize\textpm0.31} & 24.00 \\
\multicolumn{2}{l}{RKL}     & 24.11{\scriptsize\textpm0.31} & \textbf{32.09{\scriptsize\textpm0.37}} & 30.29{\scriptsize\textpm0.11} & 17.97{\scriptsize\textpm0.84} & 16.02{\scriptsize\textpm0.73} & 24.09 \\
\multicolumn{2}{l}{JS}      & 24.41{\scriptsize\textpm0.34} & 28.55{\scriptsize\textpm0.33} & 28.69{\scriptsize\textpm0.10} & 17.31{\scriptsize\textpm0.32} & 16.21{\scriptsize\textpm0.52} & 23.03 \\
\multicolumn{2}{l}{TVD}     & 24.71{\scriptsize\textpm0.74} & 29.23{\scriptsize\textpm0.25} & 29.12{\scriptsize\textpm0.05} & 16.64{\scriptsize\textpm0.83} & 16.19{\scriptsize\textpm0.63} & 23.18 \\
\multicolumn{2}{l}{SKL}     & 25.32{\scriptsize\textpm0.54} & 31.10{\scriptsize\textpm0.38} & 29.89{\scriptsize\textpm0.11} & 17.45{\scriptsize\textpm0.69} & 16.32{\scriptsize\textpm0.33} & 24.01 \\
\multicolumn{2}{l}{SRKL}    & 24.93{\scriptsize\textpm0.18} & 30.52{\scriptsize\textpm0.31} & \underline{30.62{\scriptsize\textpm0.15}} & 17.17{\scriptsize\textpm0.68} & 16.41{\scriptsize\textpm0.36} & 23.93 \\
\multicolumn{2}{l}{AKL}     & \underline{25.50{\scriptsize\textpm0.53}} & 30.41{\scriptsize\textpm0.28} & 30.55{\scriptsize\textpm0.08} & 17.52{\scriptsize\textpm0.57} & \underline{16.79{\scriptsize\textpm0.34}} & \underline{24.15} \\
\rowcolor{yellow!30}
\multicolumn{2}{l}{\textbf{ToDi (Ours)}}    & \textbf{26.26{\scriptsize\textpm0.31}} & \underline{31.53{\scriptsize\textpm0.22}} & \textbf{31.29{\scriptsize\textpm0.17}} & \underline{18.14{\scriptsize\textpm0.23}} & \textbf{16.96{\scriptsize\textpm0.23}} & \textbf{24.83} \\
\bottomrule
\end{tabular}%
}
\caption{Across all distillation settings, our proposed ToDi consistently outperforms every baseline in ROUGE-L score. The best result is shown in \textbf{bold}, and the second best is \underline{underlined}.}

\label{tab:performance_reordered}

\end{table*}

\subsection{Results}
\paragraph{Overall Performance}
We first evaluate the overall instruction-following performance of ToDi against baselines using ROUGE-L. Table~\ref{tab:performance_reordered} presents the performance of the teacher and student models under different teacher–student configurations, compared across various knowledge distillation methods. Our proposed ToDi achieves the highest average score on all five instruction-following tasks for both teacher–student pairs, outperforming all baseline methods, showing that ToDi effectively transfers the knowledge of the teacher to the student. We demonstrate that ToDi consistently outperforms all single-divergence baselines and even surpasses an approach that uses a single, global weight across the entire vocabulary. These results indicate that dynamic, token-level adjustment of divergence weights—tailored to each token’s predicted probability discrepancy—yields significant performance gains.

\begin{figure}[!h]
  \centering
  \includegraphics[width=\columnwidth]{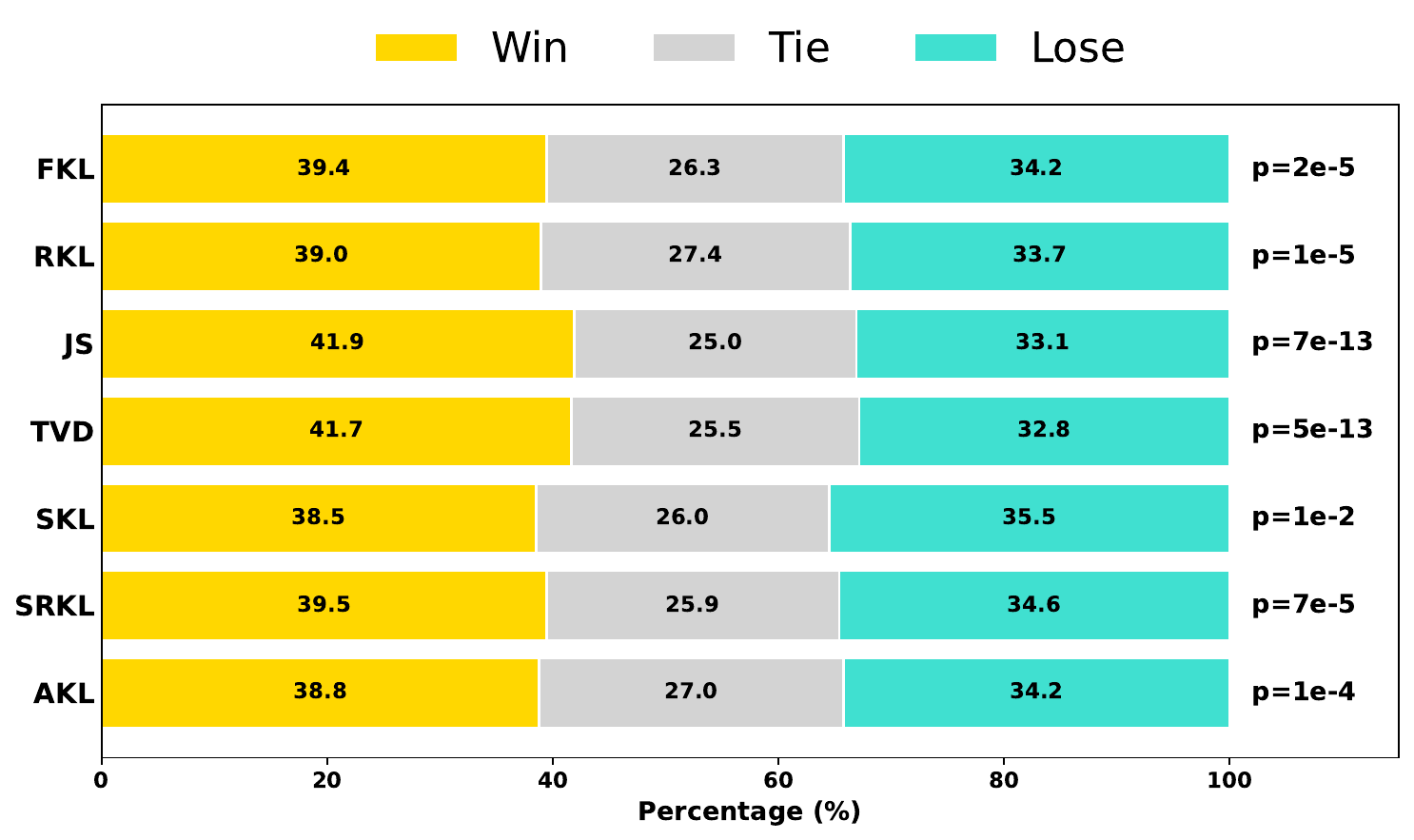}
  
\caption{GPT-4 pairwise evaluation of TinyLLaMA models trained with various KD methods on 5,000 UnNI examples. Bars show Win/Tie/Lose proportions; p-values on right.}
  \label{fig:GPT-4}
  \vspace{-3mm}
\end{figure}

\paragraph{Preference Evaluation via GPT-4}
We further evaluate ToDi through a pairwise comparison experiment using GPT-4. 
We also evaluate the subjective quality of responses generated by models trained with ToDi using a GPT-4. 
We randomly select 5,000 samples from the UnNI dataset and compare the responses generated by a TinyLLaMA model trained with ToDi to those produced by models trained with alternative divergence objectives. GPT-4 judged which response was superior. As shown in Figure~\ref{fig:GPT-4}, ToDi consistently achieved higher win rates across all comparisons. In most cases, these improvements were statistically significant (p < 0.001), confirming ToDi’s superiority over the baselines. For additional details, refer to Appendix~\ref{app:gpt4_evaluation_appendix}.

\begin{table*}[!ht]
\small
\centering
\makebox[\linewidth][c]{
\begin{tabular}{ll|ccccc}
\toprule
\multicolumn{2}{l|}{\textbf{Methods}}      & \textbf{GPT2} & \textbf{LLaMa2} & \textbf{OLMo2} & \textbf{Qwen2.5} & \textbf{Gemma3}  \\
\midrule
\rowcolor{gray!30}
\multicolumn{2}{l|}{Teacher}          & 26.66{\scriptsize\textpm0.30} & 28.88{\scriptsize\textpm0.23} & 30.24{\scriptsize\textpm0.48}& 27.42{\scriptsize\textpm0.63} & 30.60{\scriptsize\textpm0.42}\\
\multicolumn{2}{l|}{SFT}              & 23.09{\scriptsize\textpm0.53} & 23.36{\scriptsize\textpm0.26} & 24.53{\scriptsize\textpm0.41} & 24.89{\scriptsize\textpm0.25} & 24.12{\scriptsize\textpm0.37} \\
\multicolumn{2}{l|}{FKL}              & 24.06{\scriptsize\textpm0.43} & 25.40{\scriptsize\textpm0.50} & 26.88{\scriptsize\textpm0.57} & 26.71{\scriptsize\textpm0.56} & 26.88{\scriptsize\textpm0.35}\\
\multicolumn{2}{l|}{RKL}              & 24.22{\scriptsize\textpm0.18} & 24.11{\scriptsize\textpm0.31} & 25.98{\scriptsize\textpm0.46} & 27.14{\scriptsize\textpm0.32} & 28.69{\scriptsize\textpm0.14}  \\
\multicolumn{2}{l|}{JS}               & 23.77{\scriptsize\textpm0.29} & 24.41{\scriptsize\textpm0.34} & 25.39{\scriptsize\textpm0.59} & 26.82{\scriptsize\textpm0.12} & 25.10{\scriptsize\textpm0.40}  \\
\multicolumn{2}{l|}{TVD}              & 23.90{\scriptsize\textpm0.61} & 24.71{\scriptsize\textpm0.74} & 25.60{\scriptsize\textpm0.34} & 26.78{\scriptsize\textpm0.52} & 26.06{\scriptsize\textpm0.21} \\
\multicolumn{2}{l|}{SKL}              & 24.05{\scriptsize\textpm0.31} & 25.32{\scriptsize\textpm0.54} & 25.86{\scriptsize\textpm0.31} & 27.04{\scriptsize\textpm0.17} & 26.16{\scriptsize\textpm0.35} \\
\multicolumn{2}{l|}{SRKL}             & 24.20{\scriptsize\textpm0.40} & 24.93{\scriptsize\textpm0.18} & 26.03{\scriptsize\textpm0.12}& 26.74{\scriptsize\textpm0.54}  & 25.90{\scriptsize\textpm0.59} \\
\multicolumn{2}{l|}{AKL}              & 24.67{\scriptsize\textpm0.29} & 25.50{\scriptsize\textpm0.53} & 25.97{\scriptsize\textpm0.13}& 26.66{\scriptsize\textpm0.22} & 28.53{\scriptsize\textpm0.37} \\
\rowcolor{yellow!30}
\multicolumn{2}{l|}{\textbf{ToDi (Ours)}} & \textbf{24.81{\scriptsize\textpm0.62}} & \textbf{26.26{\scriptsize\textpm0.31}} & \textbf{26.94{\scriptsize\textpm0.41}}& \textbf{27.20{\scriptsize\textpm0.34}}  & \textbf{29.03{\scriptsize\textpm0.43}}  \\

\bottomrule
\end{tabular}
} 
\caption{ROUGE-L scores on the DollyEval benchmark across diverse distillation settings with varying teacher-student model pairs, including GPT2-1.5B $\rightarrow$ GPT2-120M, LLaMA2-7B $\rightarrow$ TinyLLaMA-1.1B, OLMo2-7B $\rightarrow$ OLMo2-1B, Qwen2.5-1.5B  $\rightarrow$ Qwen2.5-0.5B and Gemma3-4B $\rightarrow$ Gemma3-1B. The best result is shown in \textbf{bold}.}

\label{tab:study}
\end{table*}

\paragraph{Evaluation on Various Model Configurations}
In addition to GPT2 and LLaMa2 in Table~\ref{tab:performance_reordered}, we further evaluate ToDi’s performance across diverse teacher–student configurations. We further experiment with OLMo2~\cite{olmo20252olmo2furious} (7B -> 1B), Qwen2.5~\cite{qwen2025qwen25technicalreport} (1.5B -> 0.5B), and Gemma3~\cite{gemmateam2025gemma3technicalreport} (4B -> 1B) for the DollyEval benchmark. As shown in Table~\ref{tab:study}, ToDi consistently outperforms existing baselines under all five configurations.
This demonstrates that ToDi can transfer knowledge robustly and effectively across different teacher–student setups.

\subsection{Analysis}

\begin{figure}[!h]
  \centering
  \includegraphics[width=\columnwidth]{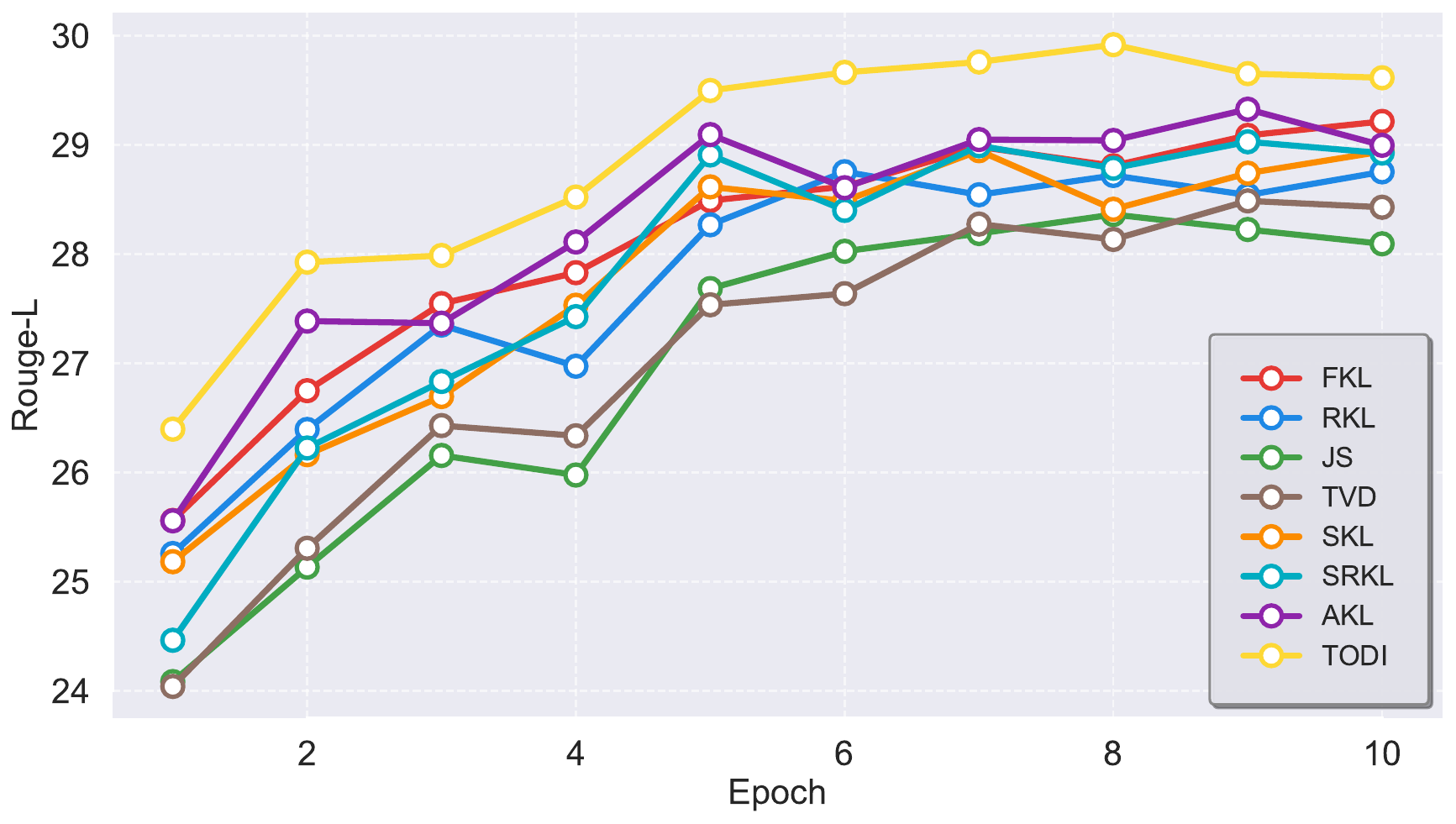}
    \caption{Validation ROUGE-L scores per epoch for TinyLLaMA using various KD methods.}
    
  \label{fig:Faster Convergence via Fine-Grained Alignment}
\end{figure}

\paragraph{Training Stability and Convergence}
We analyze the training dynamics of ToDi to assess its stability and convergence behavior. As shown in Figure~\ref{fig:Faster Convergence via Fine-Grained Alignment}, ToDi maintains a large performance margin over other methods at every epoch, achieving the highest scores throughout training. In particular, ToDi outperforms all baselines by a wide margin in the first epoch and exhibits a steady upward trajectory during the middle epochs (2–6 epochs). In the later stages (6–10 epochs), its learning curve remains smooth and converges stably without oscillation. These results indicate that ToDi not only provides a strong training signal as a KD loss function but also ensures reliable convergence.

\paragraph{Computational Efficiency}
We compare the computational complexity of ToDi with existing methods to assess its efficiency. The efficiency of ToDi is evident not only in its performance but also in its computational complexity. For instance, AKL—which dynamically adjusts the weights of FKL and RKL globally across the entire vocabulary—incurs a time complexity of \(O(V \log V)\) due to the required sorting operations. In contrast, ToDi performs computations adaptively on a per-token basis without any sorting during loss computation. As a result, it preserves linear time complexity \(O(V)\) with respect to vocabulary size, identical to both FKL and RKL.

\begin{figure*}[!ht]
  \centering
   \captionsetup{skip=1pt}
  \begin{subfigure}[t]{0.74\textwidth}
    \centering
    \includegraphics[width=\linewidth]{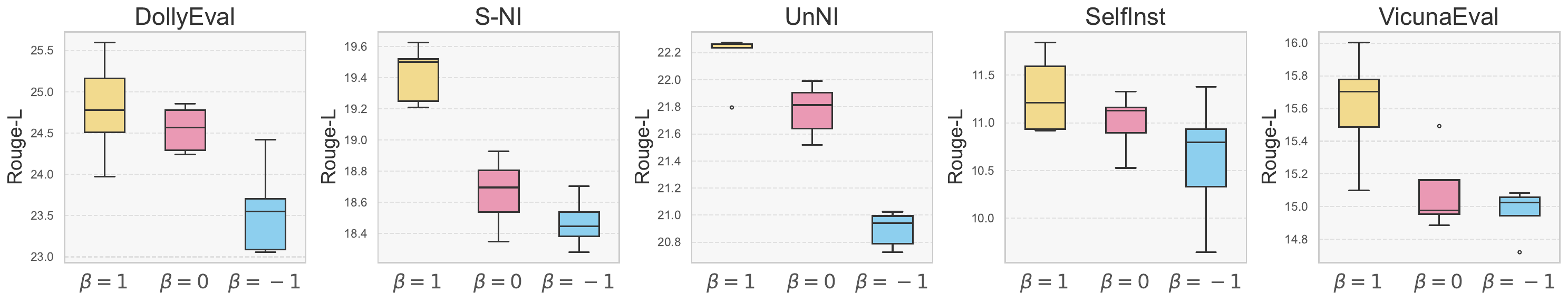}
    \phantomcaption
    \label{fig:GToDi}
  \end{subfigure}
  \hfill
  \begin{subfigure}[t]{0.24\textwidth}
    \centering
    \includegraphics[width=\linewidth]{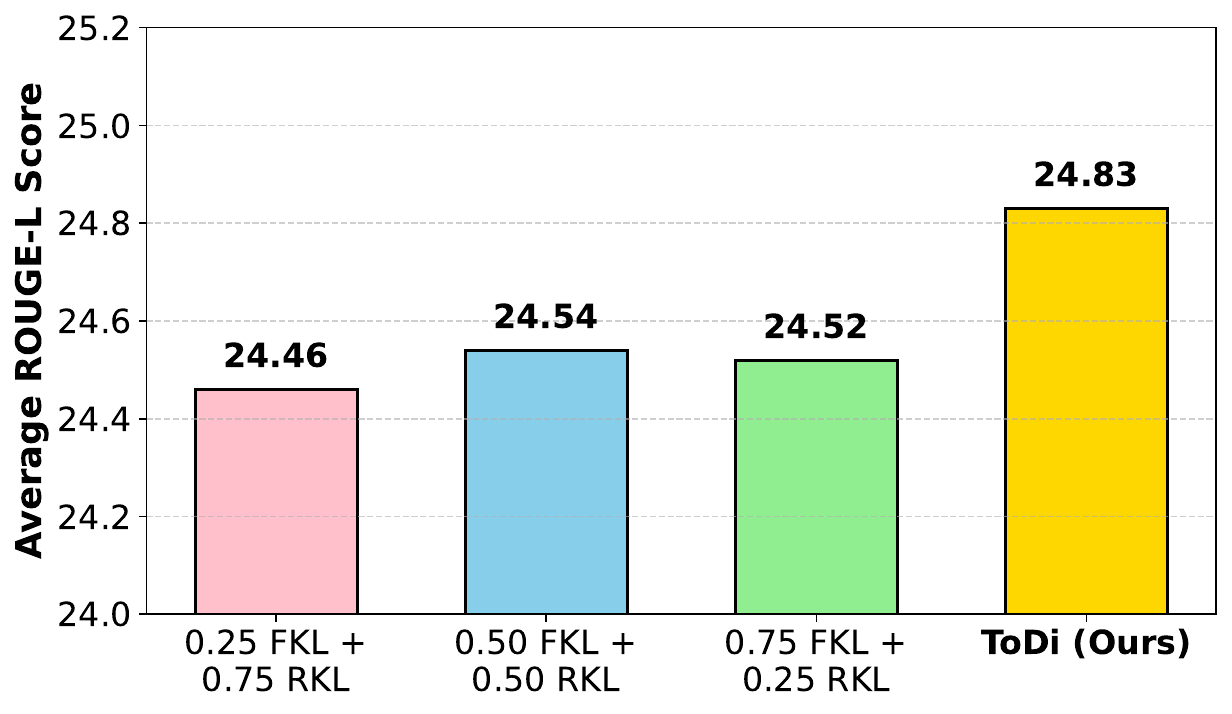}
    \phantomcaption
    \label{fig:Ablation}
  \end{subfigure}
  \caption{\textbf{(Left)} Performance comparison of Generalized ToDi with different scaling parameters $\beta \in \{1, 0, -1\}$ across five evaluation datasets. The dynamic weighting scheme ($\beta = 1$) outperforms the static setting ($\beta = 0$), while the reversed weighting ($\beta = -1$) shows clear performance degradation on all datasets. \textbf{(Right)} Average ROUGE-L scores on five instruction-following benchmarks for fixed-ratio FKL–RKL mixtures uniformly applied across the entire vocabulary distribution versus ToDi’s token-wise weighting strategy.}
  \label{fig:combined}
  \vspace{-2mm}
\end{figure*}

\paragraph{Effect of the Generalization Parameter \(\beta\)}
To analyze the impact of the scaling parameter \(\beta\), we compare the three settings \(\beta \in \{1, 0, -1\}\) in generalized ToDi. 
\(\beta = 1\) corresponds to the default ToDi configuration; \(\beta = 0\) fixes \(\alpha = 0.5\), resulting in an equal combination of FKL and RKL (i.e., Jeffreys divergence); and \(\beta = -1\) reverses the weighting direction, amplifying FKL when \(q_\theta > p\) and RKL when \(p > q_\theta\). Experimental results with GPT2-120M are shown in Figure~\ref{fig:combined} (Left). The dynamic weighting scheme (\(\beta=1\)) outperforms both the static setting (\(\beta=0\)) and the reversed setting (\(\beta=-1\)), with the reversed setting exhibiting even lower performance than the static scheme, indicating that ToDi’s adaptive weight adjustment contributes to performance improvements. 

\begin{table}[!h]
\small

\resizebox{\linewidth}{!}{%

\centering
\begin{tabular}{c|ccccc|c}
\toprule

\(\beta\) & Dolly & S-NI & UnNI  & Self & Vicuna & Average \\
\midrule
\midrule

0.6 & 24.44 & 18.17 & 22.44 & 10.88 & 16.09 & 18.40 \\
0.8 & 24.50 & 19.15 & 22.04 & 10.76 & 15.74 & 18.44 \\
1 & 24.81 & 19.42 & 22.16 & 11.30 & 15.61 & 18.66 \\
1.2 & 24.29& 18.85 & 21.86 & 11.15  & 15.69 & 18.37 \\
\midrule

\(\infty\) & 24.30& 18.96 & 21.89 & 10.93 & 15.11 & 18.24 \\

\bottomrule
\end{tabular}
}
\caption{Comparison of ROUGE-L scores of GPT-2 student models under different values of the scaling parameter \(\beta\).}
\vspace{-2mm}
\label{tab:Sensitivity Study for}
\end{table}

\paragraph{Sensitivity Analysis on \texorpdfstring{\(\beta\)}{beta}}
\label{sec:beta_sensitivity}

Table~\ref{tab:Sensitivity Study for} reports ROUGE-L scores as a function of the scaling parameter \(\beta \in \{0.6, 0.8, 1.0, 1.2, \infty\}\). The experiments show that \(\beta=1.0\) achieves the highest average score of 18.66. Two key trends are observed:

\begin{itemize}
    \item \textbf{Low-sensitivity regime (\(\beta < 1\))}: As \(\beta\) decreases, the sigmoid’s slope becomes shallower, causing the weight \(\alpha_{t,i}\) to converge toward 0.5. This nearly fixed combination of FKL and RKL reduces responsiveness to token-level prediction discrepancies, degrading training effectiveness. Indeed, at \(\beta = 0.6\), the average performance drops to 18.40.
    
    \item \textbf{High-sensitivity regime (\(\beta \rightarrow \infty\))}: As \(\beta\) grows large, the sigmoid approaches a step function and the weight \(\alpha_{t,i}\) becomes discrete:
    
    {\small 
    $\alpha_{t,i} \xrightarrow{\beta\to\infty} \mathbf{1}\bigl[p\bigl(v_i \mid \mathbf{y}_{<t}, \mathbf{x}\bigr) > q_\theta\bigl(v_i \mid \mathbf{y}_{<t}, \mathbf{x}\bigr)\bigr]$.}

    This fully separates the application of FKL and RKL, introducing discontinuities in the learning signal near the boundary \(p \approx q_\theta\). Such abrupt transitions undermine training stability, and the average performance declines to 18.24.
\end{itemize}

\paragraph{Token-wise vs. Uniform Divergence Control}

Rather than applying a fixed FKL–RKL ratio uniformly across all tokens, ToDi dynamically adjusts this balance on a per-token basis. To validate this effect, we conduct comparative experiments on a TinyLLaMA model using the fixed FKL–RKL mixtures schemes. As shown in Figure~\ref{fig:combined} (Right), ToDi consistently achieves higher ROUGE-L scores than all fixed-ratio schemes. This demonstrates that flexible, token-level ratio adjustment, rather than a uniform application across the vocabulary, is the key to performance improvements.

\begin{table}[!h]
\small
\centering
\makebox[\linewidth][c]{%
\begin{tabular}{ll|cc}
\toprule
\multicolumn{2}{l|}{\textbf{Methods}}   & \textbf{GPT2} & \textbf{TinyLLaMA}  \\
\midrule
\multicolumn{2}{l|}{AKL}  & 0.477 & 0.599\\
\multicolumn{2}{l|}{ToDi} & 0.482 & 0.610 \\

\bottomrule
\end{tabular}%
} 
\caption{Pearson similarities for AKL and ToDi using trained GPT-2 and TinyLLaMA models in Section~\ref{sec:Experiments}, with distributions computed from the \texttt{databricks/dolly-15k} training set. }
\vspace{-2mm}
\label{tab:AKL vs}
\end{table}

\paragraph{Coarse vs. Fine-Grained Weighting}
To demonstrate that a student model trained with ToDi more accurately learns the teacher distribution than one trained with AKL, we compare the distributions generated by each student model to the teacher distribution following \citet{NEURIPS2022_da669dfd}.
Table~\ref{tab:AKL vs} summarizes our analysis by reporting the Pearson similarity between the teacher and student model distributions. 
ToDi achieves higher Pearson similarity than AKL, which—despite adaptively combining forward and reverse KL at each time step—applies a uniform mixing ratio across the entire vocabulary. This indicates that ToDi’s dynamic, per-token mixing more accurately captures the teacher distribution.

\section{Conclusion}

We present ToDi, a novel token-wise distillation method that dynamically balances FKL and RKL based on per-token prediction discrepancies. Our gradient analysis shows that FKL corrects underestimation while RKL suppresses overestimation, and ToDi leverages this by using a sigmoid-based weight per token. Experiments on multiple instruction-following benchmarks demonstrate that ToDi consistently outperforms existing baselines, and GPT-4 pairwise preference evaluations confirm its superiority. Finally, we introduce a unified weighting framework and validate its effectiveness via extensive ablations.

\section*{Limitations}

ToDi precisely captures token-level prediction discrepancies between the teacher and student models, thereby enabling effective distribution alignment. However, ToDi assumes that the teacher and student share an identical vocabulary, which limits its direct applicability when the two models employ different vocabularies.
Moreover, ToDi requires access to the full token probability distribution of the teacher model, restricting its use to open-source LLMs that expose per-token logits.

Experiments on extremely large-scale models were not conducted due to computational resource constraints. Nevertheless, ToDi consistently outperforms existing methods across a diverse range of models, including GPT2-120M and TinyLLaMA-1.1B, demonstrating its practicality and efficiency.

\section*{Acknowledgments}
This work was supported by the National Research Foundation of Korea(NRF) grant funded by the Korea government(MSIT)(RS-2025-24683575).
This work was supported by the Institute of Information \& Communications Technology Planning \& Evaluation (IITP) grant funded by the Korea government (MSIT) [RS-2021-II211341, Artificial Intelligence Graduate School Program (Chung-Ang University)].
This work was supported by the ICT Credit-Linked Internship Program.
We would also like to thank Prof. Changhee Lee (Korea University) for his valuable feedback and discussions, and Yonghyun Jun and Junhyuk Choi (Undergraduate students at Chung-Ang University) for their assistance and contributions to this work.


\clearpage
\appendix

\section{Gradient Derivations}
\label{sec:appendix_gradient}

\subsection{Derivation of FKL Gradient}
\label{app:fkl_grad}

We consider the forward KL divergence term at time step $t$ and vocabulary token $v_i$, defined as:
\begin{equation}
D_{\text{FKL}}^{(t,i)}(p, q_\theta) = p_i \log \frac{p_i}{q_i}
\label{eq:app_fkl_def}
\end{equation}
where:
\begin{equation}
p_i := p(v_i \mid \mathbf{y}_{<t}, \mathbf{x}), \quad q_i := q_\theta(v_i \mid \mathbf{y}_{<t}, \mathbf{x})
\label{eq:app_fkl_defs}
\end{equation}

To compute the gradient with respect to $q_i$, we apply the product rule:
\begin{equation}
\frac{\partial}{\partial q_i} D_{\text{FKL}}^{(t,i)}(p, q_\theta)
= \frac{\partial}{\partial q_i} \left[ p_i \log \frac{p_i}{q_i} \right]
\end{equation}
Since $p_i$ is independent of $q_i$, we treat it as a constant:
\begin{equation}
= p_i \cdot \frac{\partial}{\partial q_i} \left( \log p_i - \log q_i \right)
= -p_i \cdot \frac{1}{q_i}
\end{equation}

Thus, the gradient becomes:
\begin{equation}
\frac{\partial}{\partial q_\theta(v_i \mid \mathbf{y}_{<t}, \mathbf{x})} D_{\text{FKL}}^{(t,i)}(p, q_\theta)
= -\frac{p(v_i \mid \mathbf{y}_{<t}, \mathbf{x})}{q_\theta(v_i \mid \mathbf{y}_{<t}, \mathbf{x})}
\label{eq:app_fkl_grad}
\end{equation}

\subsection{Derivation of RKL Gradient}
\label{app:rkl_grad}

We now derive the gradient for the reverse KL divergence, defined as:
\begin{equation}
D_{\text{RKL}}^{(t,i)}(p, q_\theta) = q_i \log \frac{q_i}{p_i}
\label{eq:app_rkl_def}
\end{equation}
where the same definitions apply:
\begin{equation}
p_i := p(v_i \mid \mathbf{y}_{<t}, \mathbf{x}), \quad q_i := q_\theta(v_i \mid \mathbf{y}_{<t}, \mathbf{x})
\label{eq:app_rkl_defs}
\end{equation}
Applying the product rule:
\begin{align}
\frac{\partial}{\partial q_i} D_{\text{RKL}}^{(t,i)}(p, q_\theta)
&= \frac{\partial}{\partial q_i} \left[ q_i \log \frac{q_i}{p_i} \right] \notag \\
&= \frac{\partial}{\partial q_i} \left( q_i \log q_i - q_i \log p_i \right)
\end{align}

\noindent
Since $\log p_i$ is constant w.r.t. $q_i$, the derivative simplifies to:
\begin{align}
\frac{\partial}{\partial q_i} D_{\text{RKL}}^{(t,i)}(p, q_\theta)
&= \left( \log q_i + 1 \right) - \log p_i \notag \\
&= \log \frac{q_i}{p_i} + 1
\end{align}
Hence, the final gradient expression is:
\begin{align}
\frac{\partial}{\partial q_\theta(v_i \mid \mathbf{y}_{<t}, \mathbf{x})} D_{\text{RKL}}^{(t,i)}(p, q_\theta)\\
= \log \frac{q_\theta(v_i \mid \mathbf{y}_{<t}, \mathbf{x})}{p(v_i \mid \mathbf{y}_{<t}, \mathbf{x})} \notag + 1
\end{align}

\section{Proof of Sigmoid Weight-Function Properties}
\label{app:sigmoid}

For the ToDi weight function
\begin{equation}
\alpha_{t,i} = \sigma\!\Biggl(\log \frac{p\bigl(v_i \mid \mathbf{y}_{<t}, \mathbf{x}\bigr)}{q_\theta\bigl(v_i \mid \mathbf{y}_{<t}, \mathbf{x}\bigr)}\Biggr)
\end{equation}
we prove the following:

\begin{itemize}
  \item If 
  \(
    p(v_i \mid \mathbf{y}_{<t}, \mathbf{x}) > q_\theta(v_i \mid \mathbf{y}_{<t}, \mathbf{x}),
  \)
  then
  \(
    \log\frac{p(v_i \mid \mathbf{y}_{<t}, \mathbf{x})}{q_\theta(v_i \mid \mathbf{y}_{<t}, \mathbf{x})} > 0
    \;\Rightarrow\; \alpha_{t,i} > 0.5,
  \)
  which increases the contribution of FKL.

  \item If 
  \(
    q_\theta(v_i \mid \mathbf{y}_{<t}, \mathbf{x}) > p(v_i \mid \mathbf{y}_{<t}, \mathbf{x}),
  \)
  then
  \(
    \log\frac{p(v_i \mid \mathbf{y}_{<t}, \mathbf{x})}{q_\theta(v_i \mid \mathbf{y}_{<t}, \mathbf{x})} < 0
    \;\Rightarrow\; \alpha_{t,i} < 0.5,
  \)
  which increases the contribution of RKL.

  \item Let 
  \(
    r = {p(v_i \mid \mathbf{y}_{<t}, \mathbf{x})}/{q_\theta(v_i \mid \mathbf{y}_{<t}, \mathbf{x})},
  \)
  so that 
  \(
    \alpha_{t,i} = \sigma(\log r).
  \)
  Then
  \[
    \frac{d\alpha_{t,i}}{dr}
    = \frac{\sigma(\log r)\bigl(1-\sigma(\log r)\bigr)}{r}
    > 0
  \]
  implying that \(\alpha_{t,i}\) is monotonically increasing in \(r\).

  \item Since \(\forall z,\; \sigma(z)\in(0,1)\), it follows that \(\alpha_{t,i}\in(0,1)\).
\end{itemize}

\section{Jeffreys-Inspired Weighting with Stop-Gradient}
\label{app:todi_vs_jeffreys}
The token-wise weight \(\alpha_{t,i}\) in ToDi is inspired by Jeffreys divergence. In this section, we outline this connection and, in particular, show analytically how applying a stop-gradient (\texttt{detach}) to \(\alpha_{t,i}\) yields gradients that differ from those of standard Jeffreys divergence.

At time step \(t\) for token \(v_i \in \mathcal{V}\), the Jeffreys divergence can be written using Equation~\ref{eq:fkl_def} and Equation~\ref{eq:rkl_def} as:
\begin{equation}
D_{\text{Jeffreys}}^{(t,i)}(p, q_{\theta})
\;=\;
D_{\text{FKL}}^{(t,i)}(p, q_{\theta})
+
D_{\text{RKL}}^{(t,i)}(p, q_{\theta})
\label{eq:jeffreys_token}
\end{equation}

The ToDi weighting function \(\alpha_{t,i}\) can then be derived from Jeffreys divergence as:
\begin{equation}
\begin{aligned}
&p_i \log\frac{p_i}{q_i}
\;+\;
q_i \log\frac{q_i}{p_i} \\[6pt]
&= p_i \log\frac{p_i}{q_i}
   \;-\;
   q_i \log\frac{p_i}{q_i} \\[6pt]
&= (p_i - q_i)\,
   \log\frac{p_i}{q_i} \\[6pt]
&= \frac{p_i^2 - q_i^2}{p_i + q_i}\,
   \log\frac{p_i}{q_i} \\[6pt]
&= \frac{p_i^2}{p_i + q_i}\,
   \log\frac{p_i}{q_i}
   \;-\;
   \frac{q_i^2}{p_i + q_i}\,
   \log\frac{p_i}{q_i} \\[6pt]
&= \frac{p_i^2}{p_i + q_i}\,
   \log\frac{p_i}{q_i}
   \;+\;
   \frac{q_i^2}{p_i + q_i}\
   \log\frac{q_i}{p_i} \\[6pt]
&= \frac{p_i}{p_i + q_i}\,
   \bigl(p_i \log\tfrac{p_i}{q_i}\bigr)
   \;+\;
   \frac{q_i}{p_i + q_i}\
   \bigl(q_i \log\tfrac{q_i}{p_i}\bigr)\\
&= \sigma\!\left(\log\frac{p_i}{q_i}\right)\,
  \Bigl(p_i \log\frac{p_i}{q_i}\Bigr)\\
&\quad+\;
\Bigl(1-\sigma\!\left(\log\frac{p_i}{q_i}\right)\Bigr)\,
  \Bigl(q_i \log\frac{q_i}{p_i}\Bigr)
\end{aligned}
\label{eq:jeffreys_expand_app}
\end{equation}

where, for brevity, we denote
\(
p_i \coloneqq p(v_i \mid \mathbf{y}_{<t},\mathbf{x})
\)
and
\(
q_i \coloneqq q_\theta(v_i \mid \mathbf{y}_{<t},\mathbf{x}).
\)
In ToDi, \(\sigma(\log\frac{p_i}{q_i})\) is \texttt{detached} so that no gradient flows through it. As a result, \(\alpha_{t,i}\) acts purely as a constant weight, leading to an optimization behavior that diverges from Jeffreys divergence.

To clarify this difference, we compare derivatives with respect to \(q_\theta(v_i \mid \mathbf{y}_{<t}, \mathbf{x})\):
\begin{itemize}
    \item Jeffreys divergence derivative:
    \begingroup
    \small
    \begin{equation}
    \frac{\partial}{\partial q_\theta} \left[ p \log \frac{p}{q_\theta} + q_\theta \log \frac{q_\theta}{p} \right]
    = -\frac{p}{q_\theta} + \log \frac{q_\theta}{p} + 1
    \label{eq:jeffrey_grad}
    \end{equation}
    \endgroup

    \item ToDi derivative (\(\alpha_{t,i}\) is \texttt{detached}, so treated as constant):
    \begingroup
    \small
    \begin{equation}
    \begin{aligned}
    &\frac{\partial}{\partial q_\theta} \left[ 
    \alpha_{t,i} \cdot p \log \frac{p}{q_\theta} 
    + (1 - \alpha_{t,i}) \cdot q_\theta \log \frac{q_\theta}{p} \right] \\
    &= \alpha_{t,i} \left( -\frac{p}{q_\theta} \right) 
    + (1 - \alpha_{t,i}) \left( \log \frac{q_\theta}{p} + 1 \right)
    \end{aligned}
    \label{eq:ToDi_grad}
    \end{equation}
    \endgroup
\end{itemize}

Using the \texttt{detached} weight \(\alpha_{t,i}\), ToDi increases the weight on \(D_{\text{FKL}}^{(t,i)}(p, q_\theta)\) when \(p > q_\theta\), elevating the student probability, and increases the weight on \(D_{\text{RKL}}^{(t,i)}(p, q_\theta)\) when \(q_\theta > p\), suppressing the student probability. Unlike Jeffreys divergence, which applies divergence uniformly across the vocabulary, ToDi adaptively refines divergence intensity at the token level.

\section{Experimental Details}
\label{app:Experimental Details}
\subsection{Training details}

Training was conducted based on the setup of \citet{zhang-etal-2024-dual}. For GPT2-1.5B, we employed the publicly released model from \citet{gu2024minillm}, while GPT2-120M was trained for 20 epochs with a learning rate of \(5\times10^{-4}\). The TinyLLaMA and LLaMA2 models were trained for 10 epochs with a learning rate of \(1\times10^{-3}\). All experiments were carried out on a single RTX A6000 GPU. The training loss was composed by combining the KD loss and the cross-entropy loss in equal proportions (0.5:0.5). Detailed hyperparameter settings for each model are summarized in Table~\ref{tab:finetune-settings}.


\begin{table}[t]  
\small
\centering
\resizebox{\linewidth}{!}{%

\begin{tabular}{cc|ccc}
\toprule
\multicolumn{2}{c|}{\textbf{Settings}} & \textbf{GPT2} & \textbf{TinyLLaMA} & \textbf{LLaMA2}  \\
\midrule
\multicolumn{2}{c|}{Epoch} & 20 & 10 & 10\\
\multicolumn{2}{c|}{Learning Rate} & 5e-4 & 1e-3 & 1e-3\\
\multicolumn{2}{c|}{Batch Size} & 32 & 32 & 32\\
\multicolumn{2}{c|}{Fine-Tuning Method} & Full & LoRA & LoRA\\
\multicolumn{2}{c|}{LoRA Rank} & - & 256 & 256\\
\multicolumn{2}{c|}{LoRA Alpha} & - & 8 & 8\\
\multicolumn{2}{c|}{LoRA Dropout} & - &  0.1 &  0.1\\
\bottomrule
\end{tabular}
}
\caption{Hyperparameter settings for KD.}
\label{tab:finetune-settings}
\end{table}

\subsection{Evaluation details}

All test sets were processed following \citet{gu2024minillm}. The number of samples in each test set is as follows: DollyEval contains 500 examples; S-NI includes 1,694 examples with response lengths exceeding 11 tokens; UnNI comprises 10,000 examples with response lengths exceeding 11 tokens; SelfInst has 242 examples; and VicunaEval consists of 80 examples. For response generation, we used random seeds \{10, 20, 30, 40, 50\} and report the average ROUGE-L score across these seeds.

\begin{figure*}[h]
  \includegraphics[width=1\linewidth]{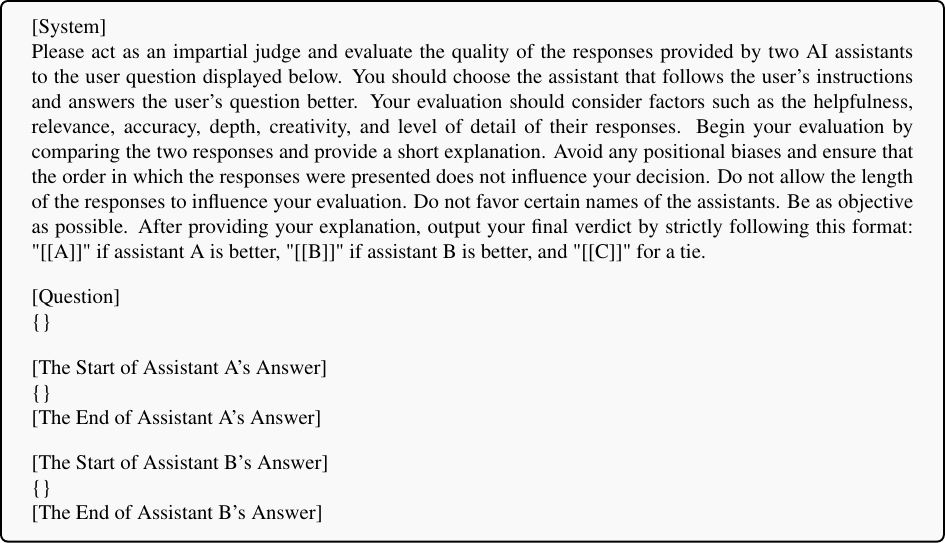}
  \caption {Prompt for GPT-4o based Evaluation.}
    \label{fig:evaluation_prompt}
\end{figure*}

\section{Details of GPT-4 Evaluation}
\label{app:gpt4_evaluation_appendix}

Pairwise comparison of model responses was performed using the \texttt{gpt-4o-2024-11-20} API, with response order randomized in the prompt to mitigate position bias. We followed the LLM-as-a-Judge evaluation protocol of \citet{zheng2023judging}, employing the pairwise comparison prompt shown in Figure~\ref{fig:evaluation_prompt}.

\end{document}